\documentclass{article}

\PassOptionsToPackage{numbers,sort&compress}{natbib}
\usepackage[preprint]{neurips_2026}
\usepackage{graphicx}
\usepackage{mathtools}
\usepackage{amssymb}
\usepackage{multirow}
\usepackage{makecell}
\usepackage{array}
\usepackage{wrapfig}
\usepackage{subcaption}

\usepackage[utf8]{inputenc} % allow utf-8 input
\usepackage[T1]{fontenc}    % use 8-bit T1 fonts
\usepackage{hyperref}       % hyperlinks
\usepackage{url}            % simple URL typesetting
\usepackage{booktabs}       % professional-quality tables
\usepackage{amsfonts}       % blackboard math symbols
\usepackage{nicefrac}       % compact symbols for 1/2, etc.
\usepackage{microtype}      % microtypography
\usepackage{xcolor}         % colors
\usepackage{bm}
\title{A Model-Internal Protocol for Assessing Multimodal Models as Integrated Systems}

\author{
  Hao Zhang\thanks{Equal contribution.} \\
  Hangzhou Institute for Advanced Study, \\
  University of Chinese Academy of Sciences \\
  \texttt{12210801@mail.sustech.edu.cn}
  \And
  Jiaxin Qi\footnotemark[1] \\
  Computer Network Information Center, \\
  Chinese Academy of Sciences \\
  \texttt{jxqi@cnic.cn}
  \AND
  Zhijiang Tang \\
  Hangzhou Institute for Advanced Study, \\
  University of Chinese Academy of Sciences \\
  \texttt{tangzhijiang24@mails.ucas.ac.cn}
  \And
  Jianqiang Huang\thanks{Corresponding author.} \\
  Computer Network Information Center, \\
  Chinese Academy of Sciences \\
  \texttt{jqhuang@cnic.cn}
}

\begin{document}

\maketitle

\begin{abstract}
  As Large Vision-Language Models increasingly aim to integrate visual generation and understanding within a single parameter space, evaluating such structural unification in a cohesive manner remains a critical challenge. Current evaluation protocols predominantly treat generative and discriminative capabilities as separate tasks, leaving a gap in system-level evaluation for unified multimodal models (UMMs). 
In this work, we propose Self-Generative-Understanding (SGU), a novel, annotation-free evaluation framework that probes the integrated capabilities of unified models through a semantic closed-loop challenge. Without requiring new annotations, SGU leverages the dual understanding-and-generation abilities of UMMs by asking them to first perceive an image and produce a textual description, subsequently reconstruct a visual context based on that description, and finally perform reasoning over the self-generated output. 
This pipeline provides a zero-cost testbed that yields an integrated performance score specifically tailored for evaluating UMMs as unified systems. Extensive experiments show that even high-performing UMMs often struggle to reason over their own generated contexts, revealing limitations that are not captured by separate evaluations of understanding or generation alone. Our work provides a complementary holistic evaluation framework and offers a foundation for benchmarking the development of next-generation unified multimodal models.
\end{abstract}
\section{Introduction}

The evolution of Artificial General Intelligence (AGI) increasingly calls for models that go beyond passive perception and can integrate multimodal understanding with visual generation~\cite{wang2025understandingworldinterveningit,durante2024agentaisurveyinghorizons}. Motivated by the intuition that the ability to generate visual content is closely related to understanding visual semantics, Multimodal Artificial Intelligence is transitioning from specialized, task-specific experts  toward Unified Multimodal Models (UMMs)~\cite{li2025girbenchversatilebenchmarkgenerating,wang2025understandingworldinterveningit,wei2025skyworkunipic20building,xie2025reconstructionalignmentimprovesunified}. By jointly training on generation and understanding objectives, UMMs consolidate discriminative reasoning and generative synthesis within a shared model, aiming to support both perception and creation in an integrated manner.

\begin{figure}[t!]
\centering\includegraphics[width=1\linewidth]{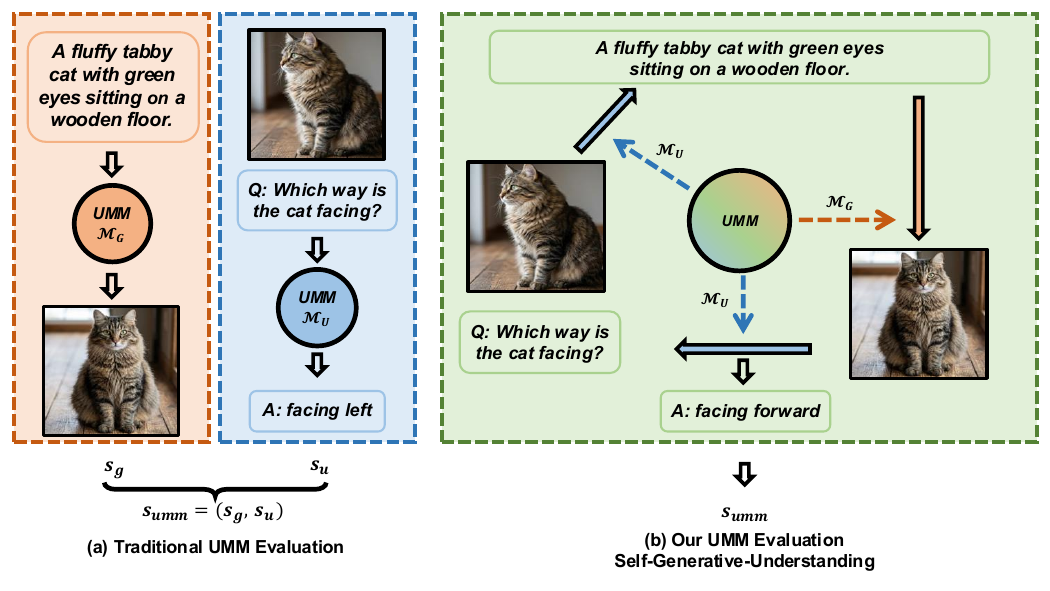}
\caption{{Illustration of traditional UMM evaluation and our Self-Generative-Understanding (SGU) framework.} 
(a) Existing evaluations often assess generation ($\mathcal{M}_G$) and understanding ($\mathcal{M}_U$) separately, producing capability-specific scores such as $s_g$ and $s_u$. These metrics remain valuable for component-wise diagnosis but do not directly provide a system-level view of how a UMM behaves when understanding and generation are jointly involved. 
(b) SGU provides a complementary closed-loop evaluation framework tailored to UMMs. It requires the model to understand an input image, generate a textual representation, reconstruct a visual context from that representation, and reason over the reconstructed image, yielding an outcome-based system-level score $s_{\text{umm}}$.}
    \label{fig:teaser}
\end{figure}

Despite the rapid architectural convergence toward unification, current evaluation protocols still primarily assess generation and understanding as separate capabilities~\cite{li2025girbenchversatilebenchmarkgenerating,liang2025roverbenchmarkingreciprocalcrossmodal}, as illustrated in Figure~\ref{fig:teaser}(a). Typically, generative ability is quantified using distributional or text-image alignment metrics such as FID~\cite{heusel2017ttur} and CLIPScore~\cite{hessel2022clipscorereferencefreeevaluationmetric}, while understanding ability is evaluated on visual question answering or multimodal reasoning benchmarks. These component-wise evaluations are important for diagnosing specific strengths and weaknesses. However, they do not directly provide a unified assessment of how well a UMM performs when understanding and generation are both involved within the same model.
This question is difficult to answer by naively aggregating heterogeneous generation and understanding metrics. For example, two UMMs may exhibit opposite capability profiles, with one stronger in generation and the other stronger in understanding. Separate scores can reveal these differences, but they do not directly indicate whether the model can function well when both abilities must interact within a unified process. This motivates a complementary evaluation perspective that treats the UMM as an integrated system rather than only as a collection of separately measured components.

To address this gap, we propose Self-Generative-Understanding (SGU), a semantic closed-loop evaluation framework explicitly designed for UMMs. SGU leverages the dual understanding-and-generation abilities of UMMs by asking the model to first perceive an input image and generate a textual description, then reconstruct a visual context based on that description, and finally answer image-grounded questions using the reconstructed image, as shown in Figure~\ref{fig:teaser}(b). Since the downstream questions are grounded in the original image, successful reasoning on the reconstructed image provides an outcome-based signal of whether the model can effectively carry out the integrated understanding-and-generation process. Any weakness in understanding, representation, or generation will be reflected in the final performance. The resulting SGU score is therefore a system-level performance score tailored to UMMs.

SGU offers three advantages. \textbf{(1) Integrated capability evaluation.} The closed-loop process activates multiple capabilities of UMMs, including image understanding, textual representation, visual generation, and downstream reasoning, without relying on external judge models or additional generation systems. \textbf{(2) System-level sensitivity.} The final score reflects the combined effect of these stages: information loss in understanding, intermediate representation, generation, or self-reasoning will reduce the final performance. Thus, SGU evaluates whether the UMM succeeds as an integrated system, while stage-wise analyses can be used for further diagnosis. \textbf{(3) Annotation-free scalability.} SGU can be instantiated on existing image-grounded reasoning benchmarks, such as MMBench~\cite{MMBench}, by reusing their original questions and ground-truth answers, requiring no new annotations.

Extensive experiments reveal a clear gap between isolated benchmark performance and integrated multimodal behavior in current UMMs. Despite strong results on individual tasks, many models suffer substantial performance degradation under the SGU loop, exposing limitations that are not captured by separate evaluations of understanding or generation alone. Stage-wise replacement experiments further help analyze which parts of the loop contribute to the degradation, while additional studies examine robustness to prompt variations, text bottlenecks, and potential shortcut behaviors. These findings show that SGU provides both a complementary system-level score and diagnostic insights for evaluating unified multimodal models.

Our contributions are summarized as follows:
\begin{itemize}
  \item We identify the need for a complementary system-level evaluation framework for UMMs, beyond separate generation and understanding metrics.
  \item We propose SGU, a semantic closed-loop evaluation framework tailored for UMMs, providing an annotation-free and model-internal system-level assessment of integrated understanding-and-generation behavior.
  \item We conduct extensive experiments and in-depth analyses of representative UMMs, providing quantitative evaluations and diagnostic insights for assessing unified multimodal behavior.
\end{itemize}   
\section{Related Work}
\label{sec:related_work}

\subsection{Unified Multimodal Models}

Multimodal intelligence has rapidly evolved from task-specific systems toward Unified Multimodal Models (UMMs), which aim to integrate visual understanding and generation within a single model~\cite{li2025girbenchversatilebenchmarkgenerating,liang2025roverbenchmarkingreciprocalcrossmodal,wei2026ggbenchgeometricgenerativereasoning}. Existing UMMs can be broadly categorized into several representative paradigms. The first line of work adopts autoregressive modeling with unified or discrete visual tokenization, allowing text and image tokens to be processed within a shared sequence modeling framework. Representative examples include Chameleon~\cite{chameleonteam2025chameleonmixedmodalearlyfusionfoundation}, Emu3~\cite{wang2024emu3nexttokenpredictionneed}, Janus-Pro~\cite{chen2025janusprounifiedmultimodalunderstanding}, and X-Omni~\cite{xomni2025}. More recent models such as Show-o2~\cite{xie2025showo2improvednativeunified} further incorporate improved tokenization or scheduling strategies to enhance the balance between understanding and generation.
The second paradigm integrates diffusion or flow-based generation modules with multimodal reasoning backbones, aiming to improve image generation quality while preserving understanding capabilities~\cite{lipman2023flowmatchinggenerativemodeling,zhou2024transfusionpredicttokendiffuse,liang2025roverbenchmarkingreciprocalcrossmodal}. Representative models such as BLIP3-o~\cite{chen2025blip3ofamilyfullyopen}, BAGEL~\cite{deng2025emergingpropertiesunifiedmultimodal}, UniWorld~\cite{lin2025uniworldv1highresolutionsemanticencoders}, and OmniGen2~\cite{wu2025omnigen2explorationadvancedmultimodal} leverage large-scale multimodal training or interleaved data to support both visual reasoning and generation.
Another line of work maintains a strong reasoning backbone while connecting it to image generation modules through learnable interfaces or routing mechanisms~\cite{pan2025transfermodalitiesmetaqueries}. This design can preserve language and vision reasoning ability while enabling generation through specialized components. In parallel, proprietary systems such as GPT-Image~\cite{openai_image_generation_api_2025b} and Gemini 2.5 Flash Image~\cite{deepmind_gemini_flash_image_2025} further demonstrate the rapid progress of unified understanding-generation systems.

\subsection{Benchmarks for Unified Multimodal Models}

Evaluation of UMMs remains challenging because current methodologies typically assess multimodal understanding and generation separately. Understanding capabilities are generally evaluated through question-answering (QA) benchmarks, such as MME~\cite{fu2025mmecomprehensiveevaluationbenchmark} and MMMU~\cite{yue2024mmmumassivemultidisciplinemultimodal, yue2025mmmu}. Conversely, generation performance often relies on reference-based benchmarks such as GenEval~\cite{ghosh2023genevalobjectfocusedframeworkevaluating}. Recent advancements, such as GIRBench~\cite{li2025girbenchversatilebenchmarkgenerating}, have extended this scope to visual editing, while VQAScore~\cite{VQAScore} leverages QA accuracy as a proxy for evaluating image generation quality. For unified evaluation, UmniBench~\cite{liu2025umnibenchunifiedunderstandgeneration} assesses generation and editing performance by utilizing the model's intrinsic understanding.

These benchmarks provide valuable component-level insights. However, they do not directly provide a system-level evaluation of how understanding and generation interact within a UMM. In addition, existing approaches often rely on references or external model-based judges, which may introduce additional biases. Building on existing VQA-style benchmarks, we propose SGU as a complementary semantic closed-loop framework for evaluating UMMs as integrated systems, without requiring new annotations or third-party judge models.
\section{Method}

\subsection{Preliminaries}

\noindent\textbf{Unified Multimodal Model (UMM).}
Formally, we define a UMM $\mathcal{M}$ as a probabilistic model designed to process and generate visual and textual modalities within a unified model. While current UMMs often share a common parameter space, their functional capabilities can be viewed through conditional projections targeting different output modalities~\cite{wang2025understandingworldinterveningit,wei2025skyworkunipic20building}. We denote the visual generation function as $\mathcal{M}_G$ and the understanding function as $\mathcal{M}_U$.

\noindent\textbf{UMM evaluation.}
The core challenge in UMM evaluation is to define a scoring function $\mathcal{S}$ that can assess the model as an integrated system. Given a multimodal dataset $\mathcal{D}=\{(v,t)\}$, where $v$ denotes visual input and $t$ denotes textual input, a system-level score can be formalized as
\begin{equation}
s_{\text{umm}} = \mathcal{S}(\mathcal{M}, \mathcal{D}).
\label{ummscore_base}
\end{equation}
Such a score should satisfy two desiderata. (1) \textit{Unified assessment}. The score should provide a scalar signal for comparing different UMMs at the system level, complementing component-wise evaluations. (2) \textit{Integrated sensitivity}. The score should reflect the combined effect of understanding and generation, such that weakness in either capability, or in their interaction within the model, can affect the final performance.

\noindent\textbf{Current evaluation protocols.}
Existing evaluations commonly assess generation and understanding using separate metrics. Formally, given a generation benchmark $\mathcal{D}_G$ and an understanding benchmark $\mathcal{D}_U$, current methods can be viewed as producing a tuple of heterogeneous scores:
\begin{equation}
s_{\text{umm}}=(s_g,s_u), \quad
s_g=\mathcal{S}_G(\mathcal{M}_G,\mathcal{D}_G), \quad
s_u=\mathcal{S}_U(\mathcal{M}_U,\mathcal{D}_U),
\label{eq:current_eval}
\end{equation}
where $\mathcal{S}_G$ denotes generation-oriented metrics such as FID~\cite{heusel2017ttur} and CLIPScore~\cite{hessel2022clipscorereferencefreeevaluationmetric}, and $\mathcal{S}_U$ denotes understanding-oriented metrics such as VQA accuracy~\cite{VQAScore}.

These component-wise evaluations are valuable for diagnosing specific capabilities. At the same time, their heterogeneous nature makes direct aggregation non-trivial. This motivates a complementary system-level perspective for understanding how a UMM behaves when understanding and generation are jointly involved.

\subsection{Our Evaluation: Self-Generative-Understanding}

\begin{figure*}[t!]
    \centering
    \includegraphics[width=1\linewidth]{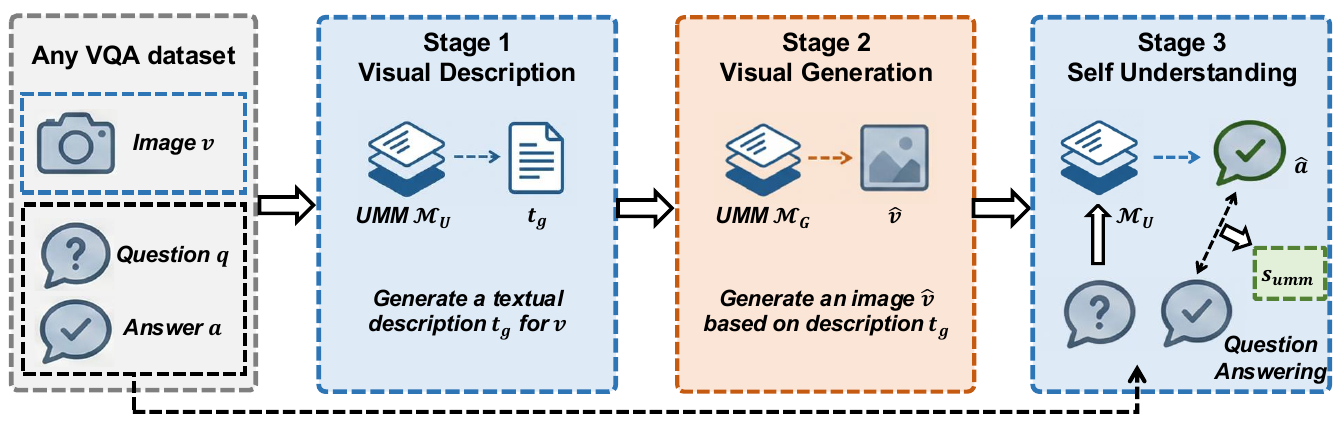}
    \caption{Overview of the Self-Generative-Understanding (SGU) protocol. Given a VQA triplet $(v, q, a)$, the UMM first generates a textual description $t_g$ from the original image $v$, reconstructs a visual context $\hat{v}$ from $t_g$, and then answers the original question $q$ based on $\hat{v}$ to obtain the SGU score $s_{\text{umm}}$. Each stage is executed in an isolated stateless session, with no memory or intermediate features carried across stages.}
    \label{fig:sgu_protocol}
\end{figure*}

We propose Self-Generative-Understanding (SGU), an annotation-free semantic closed-loop framework for system-level evaluation of UMMs. SGU is designed to evaluate whether a UMM can perform understanding and generation in an integrated manner by asking the model to operate on its own intermediate outputs.
Given a data sample $(v,t)\in\mathcal{D}$, SGU enforces a three-stage process: (1) \textit{Visual description}. The understanding function $\mathcal{M}_U$ perceives the original visual input $v$ and generates a textual description $t_g$. (2) \textit{Visual generation}. Based solely on the self-generated description $t_g$, the generation function $\mathcal{M}_G$ synthesizes a proxy visual context $\hat{v}$. (3) \textit{Self understanding}. The model then uses $\mathcal{M}_U$ to resolve the original textual task on the synthesized image $\hat{v}$ and produces the final response $\hat{t}$. The process can be formulated as
\begin{equation}
    t_g = \mathcal{M}_U(v), \quad
    \hat{v} = \mathcal{M}_G(t_g), \quad
    \hat{t} = \mathcal{M}_U(\hat{v}, t).
    \label{main_process}
\end{equation}

\subsection{SGU score}

To instantiate SGU, we adopt VQA-style settings, where each question is grounded in the original image and paired with a ground-truth answer. A data sample is formalized as a triplet $(v,q,a)$, consisting of an image $v$, a question $q$, and a ground-truth answer $a$.
Based on Eq.~\eqref{main_process}, the SGU score is calculated as
\begin{equation}
s_{\text{umm}} =
\mathbb{E}_{(v,q,a)\sim\mathcal{D}}
\left[
\mathbb{I}\left(
\operatorname{Match}
\big(
\mathcal{M}_U(\mathcal{M}_G(\mathcal{M}_U(v)), q), a
\big)
\right)
\right],
\label{eq:sgu_score}
\end{equation}
where $\mathbb{I}(\cdot)$ is the logical indicator function and $\operatorname{Match}(\cdot)$ denotes answer matching. For multiple-choice questions, $\operatorname{Match}$ checks whether the predicted option matches the ground truth. For open-form questions with deterministic answers, it applies parsing and normalization, such as case normalization, punctuation removal, and numerical matching, before comparison. This avoids relying on naive strict string matching for free-form outputs.

\noindent\textbf{What does SGU measure?}
The SGU score is an outcome-based system-level evaluation of whether a UMM can effectively operate through its own understanding-and-generation process. Since each question and answer are grounded in the original image, successful answering on the reconstructed image reflects the model's integrated multimodal capability under this process. Thus, task accuracy serves as an operational proxy for evaluating the overall success of the SGU loop. As the score aggregates the effects of different stages, stage-wise analyses can further help interpret the source of performance changes, as discussed in Section~\ref{sec:exp}.

\noindent\textbf{Relative SGU score.}
Since $s_{\text{umm}}$ is influenced by the model's base understanding ability, we introduce a relative SGU score normalized by the base understanding score $s_{\text{base}}$, which measures VQA performance on the original images and serves as a model-specific upper-bound reference:
\begin{equation}
s_{\text{base}} =
\mathbb{E}_{(v,q,a)\sim\mathcal{D}}
\!\left[
\mathbb{I}\!\left(
\operatorname{Match}\big(\mathcal{M}_U(v,q),a\big)
\right)
\right],
\quad
s_{\text{umm,r}} = \frac{s_{\text{umm}}}{s_{\text{base}}}.
\label{eq:summr}
\end{equation}
The relative score reflects how much performance is preserved after the model is required to operate through its own understanding-and-generation loop.
\section{Experiments}
\label{sec:exp}
\subsection{Implementation}

\noindent\textbf{Datasets}.
We evaluate SGU on a diverse set of representative multimodal benchmarks covering general reasoning, perception, mathematical reasoning, and text-centric visual understanding. These datasets allow us to examine the system-level behavior of UMMs across different VQA-style tasks.
\begin{itemize}
    \item \textbf{MMStar}~\cite{MMStar}, a comprehensive benchmark designed to assess general multimodal reasoning across diverse visual and textual scenarios.
    \item \textbf{MMBench}~\cite{MMBench}, which focuses on evaluating core vision--language understanding abilities through carefully curated multiple-choice questions.
    \item \textbf{MathVista}~\cite{MathVista}, targeting mathematical and logical reasoning grounded in visual contexts.
    \item \textbf{OCR-VQA}~\cite{OCR-VQA}, emphasizing text recognition and semantic understanding in visually rich, text-heavy images.
\end{itemize}

\noindent\textbf{UMMs}.
We evaluate six representative unified multimodal models across diverse architectures to analyze their system-level behavior under the SGU protocol.
\begin{itemize}
    \item \textbf{Janus-Pro-7B}~\cite{chen2025janusprounifiedmultimodalunderstanding} adopts an autoregressive framework that decouples visual encoding to mitigate conflicts between multimodal understanding and generation.
    
    \item \textbf{BAGEL-7B}~\cite{deng2025emergingpropertiesunifiedmultimodal} is a decoder-only foundation model trained on massive interleaved image-text data, subsequently exhibiting strong emergent multimodal reasoning capabilities across benchmarks.

    \item \textbf{UniWorld-V1}~\cite{lin2025uniworldv1highresolutionsemanticencoders} is a unified world model that learns consistent spatiotemporal representations for both perception and synthesis within a shared and continuous latent space.
    
    \item \textbf{Show-o2-7B}~\cite{xie2025showo2improvednativeunified} integrates discrete diffusion scheduling with autoregressive modeling to enhance generation fidelity while maintaining semantic understanding.
    
    \item \textbf{OmniGen2}~\cite{wu2025omnigen2explorationadvancedmultimodal} supports arbitrary multimodal input-output sequences with a focus on generalized instruction following in complex scenarios.
    
    \item \textbf{Ovis-U1-3B}~\cite{wang2025ovisu1technicalreport} is a lightweight unified model that enables evaluation at a smaller parameter count with lower inference cost.
\end{itemize}

\noindent\textbf{Evaluation setup}. 
For each dataset sample $(v, q, a)$, we follow the SGU protocol by first feeding the input image $v$ into the unified model to generate a textual description $t_g$, then reconstructing an image $\hat{v}$ from $t_g$, and finally performing visual question answering on $(\hat{v}, q)$ to obtain the predicted answer $\hat{t}$. The SGU score $s_{\text{umm}}$ is computed as the average accuracy over all samples.

To ensure fair and leakage-free evaluation, all stages are executed in a \emph{stateless} manner: the same unified model is reused across stages, but each stage runs in an isolated session with no memory or intermediate features preserved across steps. Throughout the evaluation, no external judge models or human annotators are introduced; all stages are performed solely by the tested unified model, resulting in a model-internal system-level evaluation protocol.

\subsection{Results Analysis}

\begin{table*}[t]
\caption{Results of various unified multimodal models on diverse VQA benchmarks under the SGU protocol. For each dataset, we report direct VQA accuracy on the original image ($s_{\text{base}}$) and SGU score ($s_{\text{umm}}$), where $s_{\text{base}}$ serves as a model-specific upper-bound reference for interpreting $s_{\text{umm}}$.}
\label{tab:main_results}
\centering
\small
\setlength{\tabcolsep}{4.2pt}
\renewcommand{\arraystretch}{1.1}
\begin{tabular}{l *{10}{c}}
\toprule
\multirow{3}{*}{\textbf{Model}}
& \multicolumn{2}{c}{\textbf{MMStar}}
& \multicolumn{2}{c}{\textbf{MMBench}}
& \multicolumn{2}{c}{\textbf{MathVista}}
& \multicolumn{2}{c}{\textbf{OCR-VQA}}
& \multicolumn{2}{c}{\textbf{Avg}} \\
\cmidrule(lr){2-3}
\cmidrule(lr){4-5}
\cmidrule(lr){6-7}
\cmidrule(lr){8-9}
\cmidrule(lr){10-11}
& Original & SGU
& Original & SGU
& Original & SGU
& Original & SGU
& Original & SGU \\
& $s_{\text{base}}$ & $s_{\text{umm}}$
& $s_{\text{base}}$ & $s_{\text{umm}}$
& $s_{\text{base}}$ & $s_{\text{umm}}$
& $s_{\text{base}}$ & $s_{\text{umm}}$
& $s_{\text{base}}$ & $s_{\text{umm}}$ \\
\midrule
\textsc{Upper Bound}
& \multicolumn{2}{c}{66.47}
& \multicolumn{2}{c}{86.13}
& \multicolumn{2}{c}{70.60}
& \multicolumn{2}{c}{81.33}
& \multicolumn{2}{c}{76.13} \\
\midrule
UniWorld-V1
& 60.67 & 38.33
& 86.13 & 71.97
& 67.30 & 37.40
& 81.33 & 28.67
& 73.86 & 44.09 \\

Janus-Pro-7B
& 47.80 & 39.27
& 76.86 & 68.20
& 41.30 & 32.50
& 69.27 & 37.45
& 58.82 & 44.36 \\

Show-o2-7B
& 55.13 & \underline{43.00}
& 83.92 & 73.95
& 50.70 & 38.90
& 63.43 & 30.34
& 63.30 & 46.55 \\

Ovis-U1
& 60.60 & \textbf{43.07}
& 83.71 & \underline{75.10}
& 68.90 & \textbf{41.80}
& 79.00 & 34.71
& 73.05 & 48.67 \\

BAGEL-7B
& 66.47 & 42.67
& 85.54 & \textbf{76.38}
& 70.60 & 39.50
& 74.55 & \underline{53.35}
& 74.29 & \underline{52.98} \\

OmniGen2
& 54.93 & \textbf{43.07}
& 82.84 & 73.66
& 63.50 & \underline{40.40}
& 79.39 & \textbf{56.59}
& 70.17 & \textbf{53.43} \\

\bottomrule
\end{tabular}
\vspace{-0.5em}
\end{table*}

\noindent\textbf{Q1. \textit{Does SGU provide a system-level view beyond isolated evaluations?}}

\noindent\textbf{A1.}
Table~\ref{tab:main_results} shows a consistent drop from direct VQA accuracy on the original image ($s_{\text{base}}$) to the SGU score ($s_{\text{umm}}$). This drop reflects the additional challenge introduced when a UMM must operate through its own understanding-and-generation process.
Importantly, models with similar $s_{\text{base}}$ can exhibit very different SGU scores. For example, on OCR-VQA, UniWorld-V1 achieves $s_{\text{base}}{=}81.33$ but drops to $s_{\text{umm}}{=}28.67$, while OmniGen2 has a comparable $s_{\text{base}}{=}79.39$ but retains a much higher $s_{\text{umm}}{=}56.59$. This indicates that isolated understanding performance alone does not fully characterize how well a UMM functions when understanding and generation are jointly involved.

\begin{wrapfigure}{r}{0.5\linewidth}

\centering

\includegraphics[width=\linewidth]{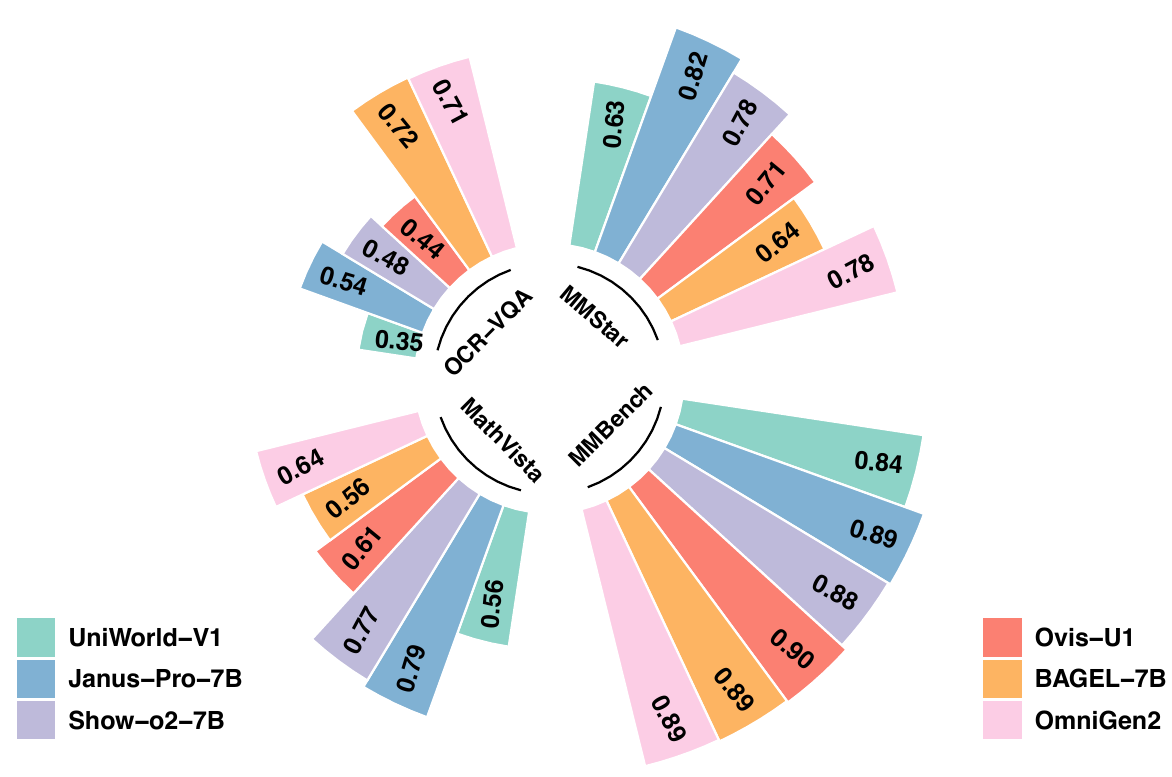}

\caption{Circular bar plot of $s_{\text{umm,r}}$.}

\label{fig:circlefig}

\end{wrapfigure}

The normalized ratio $s_{\text{umm,r}}$, defined in Eq.~\eqref{eq:summr} and visualized in Figure~\ref{fig:circlefig}, further provides a model-specific view of how much performance is preserved under the SGU loop. Datasets such as MathVista and OCR-VQA generally exhibit larger drops, suggesting that visually grounded reasoning and text-centric perception are more challenging to maintain through the integrated process. Overall, Table~\ref{tab:main_results} and Figure~\ref{fig:circlefig} show that SGU complements component-wise evaluation by providing a system-level view of UMM behavior.

\begin{figure}[!t]
\centering
\begin{minipage}[t]{0.48\linewidth}
    \centering
    \includegraphics[width=\linewidth]{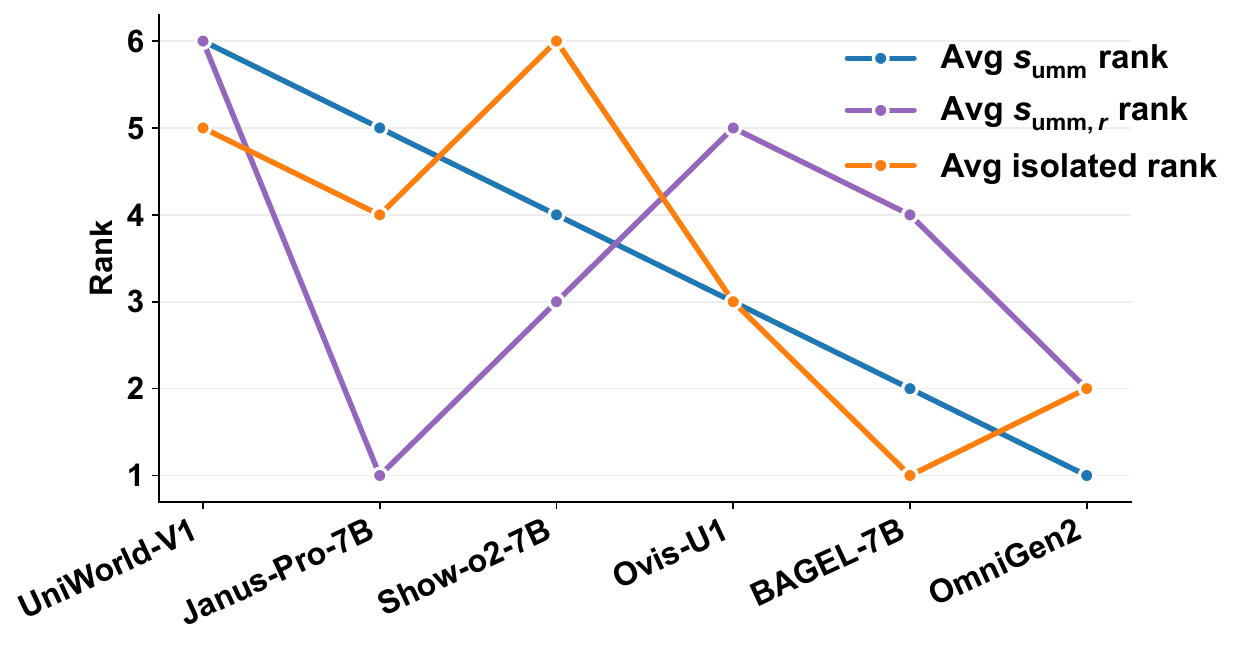}
    \captionof{figure}{Comparison of ranking performance across average SGU score $s_{\text{umm}}$, average normalized ratio $s_{\text{umm,r}}$, and an isolated average aggregated from \textit{CLIP-T}, \textit{CLIP-I}, and $s_{\text{base}}$.}
    \label{fig:ranking:trend}
\end{minipage}
\hfill
\begin{minipage}[t]{0.48\linewidth}
    \centering
    \includegraphics[width=\linewidth]{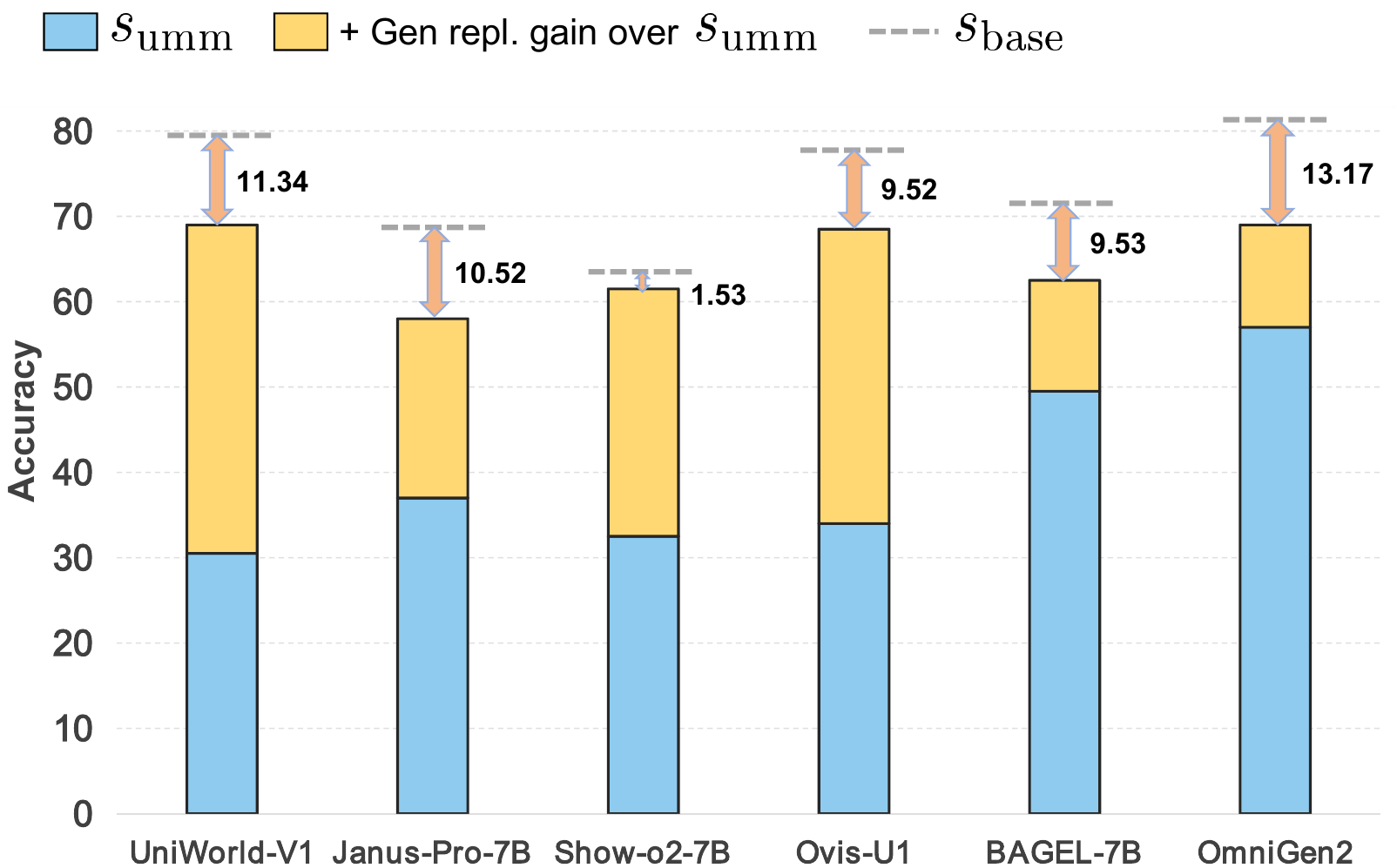}
    \captionof{figure}{Visualization of the baseline SGU accuracy $s_{\text{umm}}$, the gain from replacing the generation stage, and the residual gap to the direct VQA reference score $s_{\text{base}}$, evaluated on our constructed OCR-VQA subset.}
    \label{fig:residual_gap}
\end{minipage}
\end{figure}

To further compare SGU with conventional evaluation signals, we compute an isolated average score by aggregating commonly used component-level indicators, including direct VQA accuracy $s_{\text{base}}$ and CLIPScore~\cite{hessel2022clipscorereferencefreeevaluationmetric} between the original image and the reconstructed image or generated description. As shown in Figure~\ref{fig:ranking:trend}, the resulting ranking trend is broadly consistent with SGU. This suggests that SGU captures capability patterns reflected by existing metrics, while providing a single outcome-based score for evaluating the integrated understanding-and-generation process.

\begin{table}[t]
\centering
\begin{minipage}[t]{0.49\linewidth}
\centering
\captionof{table}{Stage-wise replacement effects on SGU score ($\Delta s_{\text{umm}}$: replaced $-$ baseline) on the constructed subsets. MV: MathVista; OCR: OCR-VQA.}
\label{tab:stage_replace_delta}
\small
\setlength{\tabcolsep}{3.5pt}
\renewcommand{\arraystretch}{1.05}
\begin{tabular}{lrrrr}
\toprule
& \multicolumn{2}{c}{Understand repl.}
& \multicolumn{2}{c}{Generation repl.} \\
\cmidrule(lr){2-3}\cmidrule(lr){4-5}
Model & MV & OCR & MV & OCR \\
\midrule
UniWorld-V1  & -4.16 & -2.52 & 13.64 & 38.89 \\
Janus-Pro-7B &  1.88 & -3.03 &  3.54 & 21.21 \\
Show-o2-7B   & -1.74 &  1.01 &  0.00 & 29.30 \\
Ovis-U1-3B   &  1.51 &  2.02 &  6.57 & 34.85 \\
BAGEL-7B     & -1.60 & -0.50 &  5.05 & 13.14 \\
OmniGen2     &  3.93 & -1.51 & 11.61 & 12.12 \\
\bottomrule
\end{tabular}
\end{minipage}
\hfill
\begin{minipage}[t]{0.47\linewidth}
\centering
\captionof{table}{Prompt-sensitivity ablation under SGU on the constructed MMBench subset with OmniGen2. We vary one stage at a time and report the mean SGU score across two runs. \textit{Und.}/\textit{Gen. Prompt} denote stage prompts; Var-1/Var-2 denote alternative templates.}
\label{tab:prompt_sens}
\small
\setlength{\tabcolsep}{3.5pt}
\renewcommand{\arraystretch}{1.05}
\begin{tabular}{lccc}
\toprule
Setting & Und. prompt & Gen. prompt & $s_{\text{umm}}$ \\
\midrule
Default & Default & Default & 73.74 \\
\midrule
Var-1 (Und) & Var1 & Default & 73.48 \\
Var-2 (Und) & Var2 & Default & 73.99 \\
Var-1 (Gen) & Default & Var1 & 75.51 \\
Var-2 (Gen) & Default & Var2 & 73.48 \\
\bottomrule
\end{tabular}
\end{minipage}
\end{table}

\noindent\textbf{Q2. \textit{How does SGU reflect and diagnose the joint effects of understanding and generation?}}

\noindent\textbf{A2.}
The SGU loop involves three key components: image-to-text understanding, image generation, and final VQA reasoning. The final VQA reasoning ability is reflected by the direct VQA reference score $s_{\text{base}}$, while the effects of the first two stages are further analyzed through stage-wise replacement experiments. Specifically, we replace the image-to-text understanding stage with Qwen3-VL-8B~\cite{qwen3vl}, and replace the image-generation stage with Qwen-Image-2512~\cite{qwenimage}, while keeping the rest of the loop unchanged. We focus on MathVista and OCR-VQA, where SGU gaps are most pronounced, and use fixed-size stratified subsets that preserve the original question-type distribution, with the same split reused for baseline and replacement runs; details are provided in Appendix~\ref{app:subset}.

Table~\ref{tab:stage_replace_delta} shows that both interventions change SGU performance, confirming that the score is affected by both understanding and generation. Replacing the generation stage yields larger positive gains across models, suggesting that visual reconstruction is a major bottleneck in the current SGU loop. However, this does not mean that SGU only measures generation: if the image-to-text stage fails to extract sufficient visual information, a stronger generator cannot recover what is missing. Consistently, understanding replacement also affects the score, though sometimes with smaller or even negative changes, suggesting that stronger isolated understanding does not always produce intermediate descriptions that are more compatible with a model's own generation and final reasoning behavior.

We further verify the role of the understanding-side intermediate representation by directly performing QA on generated captions without the image-generation stage. On OCR-VQA with OmniGen2, the score drops from $79.39$ using the original image to $60.05$ using captions alone, and further to $56.59$ under the full SGU loop. This shows that imperfect information extraction in the captioning stage already degrades performance, while generation introduces additional challenges.
Moreover, even after replacing the generation stage, most models retain a residual gap to the direct-VQA reference $s_{\text{base}}$, as shown in Figure~\ref{fig:residual_gap}. Together, these results show that SGU provides a holistic performance signal, while stage-wise and caption-only analyses help explain how understanding and generation jointly shape the final score.

\begin{figure*}[!t]
    \centering
    \includegraphics[width=0.9\linewidth]{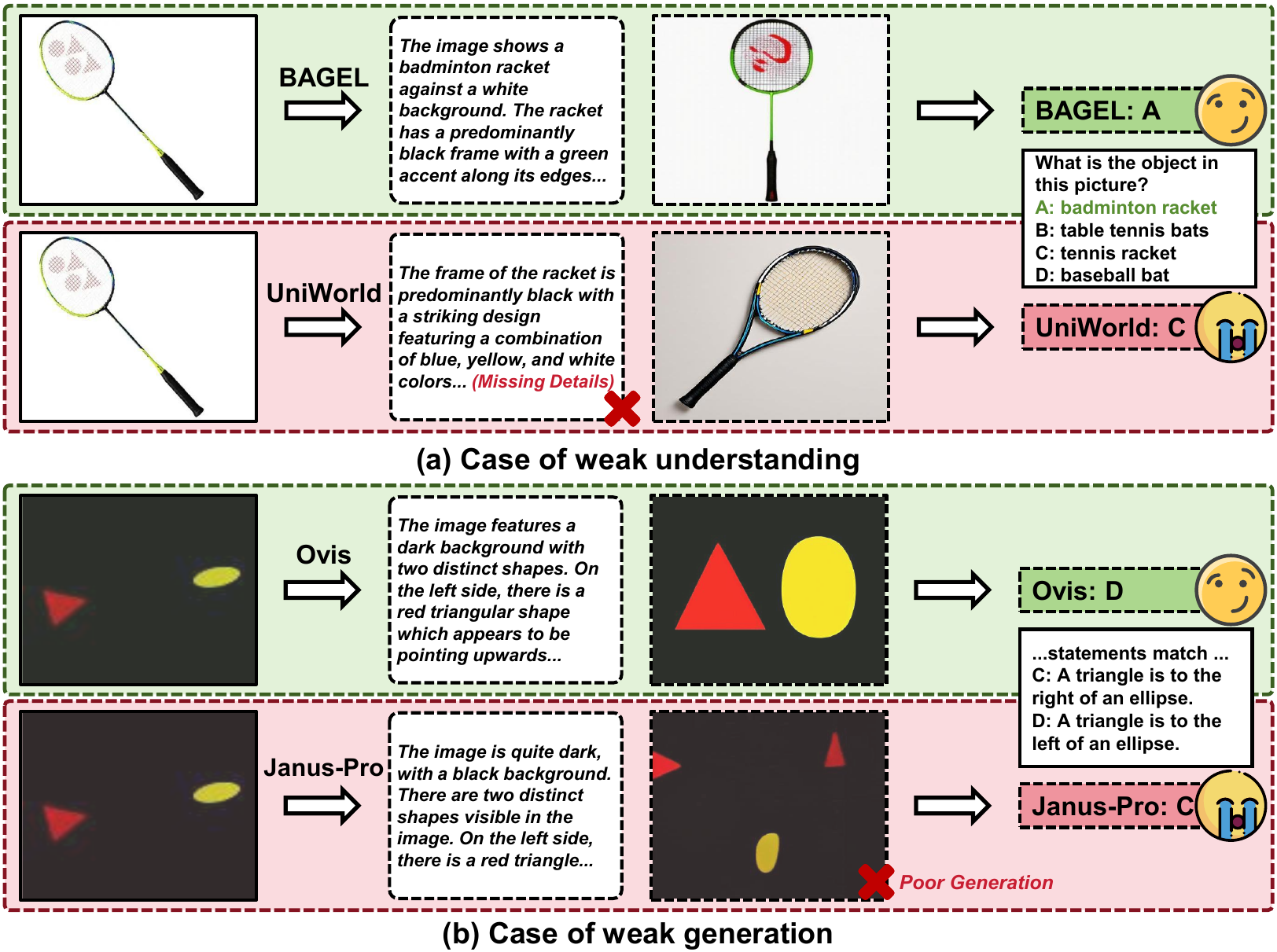}
    \caption{Qualitative case studies under the SGU protocol. The examples illustrate different loop outcomes: one model successfully performs both  understanding and generation, leading to the correct answer, while another produces an incorrect answer when visual information is incompletely extracted (Case~a) or the reconstructed image deviates from the intended content (Case~b).}
    \label{fig:case_studies}
\end{figure*}

\noindent\textbf{Q3. \textit{Is SGU sensitive to understanding and generation prompts?}}

\noindent\textbf{A3.}
We assess prompt sensitivity by rephrasing the internal prompts used in the understanding or generation stage, while keeping the rest of the SGU loop unchanged for controlled comparison. We consider three understanding prompts:
\begin{itemize}
    \item DEFAULT: \textit{Describe this image in detail}
    \item VAR1: \textit{Generate a detailed description of the visual content in this image.}
    \item VAR2: \textit{Analyze the image and provide a thorough description capturing all key elements.}
\end{itemize}
We also consider three generation prompts: 
\begin{itemize}
    \item DEFAULT: \textit{Generate an image based on the following description: \{description\}.}
    \item VAR1: \textit{Using the model-generated description directly as the generation input}
    \item VAR2: \textit{Generate a high-quality, detailed image of \{description\}.}
\end{itemize}

We change only one stage at a time, resulting in four settings, and run each setting twice with different random seeds, reporting the average accuracy.
Table~\ref{tab:prompt_sens} reports the results on OmniGen2 with our constructed MMBench subset, where all prompt variants stay close to the default score, suggesting that SGU is insensitive primarily to reasonable prompt choices within the loop and provides a stable signal under such perturbations.

\begin{table}[t]
\caption{Image-swap check on the constructed MMBench subset. We pair models with similar direct VQA accuracy $s_{\text{base}}$. For each target model $A$ (tested for potential shortcut exploitation), $s_{\text{umm}}=\mathrm{Acc}(A\!\to\!A)$ evaluates $A$ on images generated by itself, while $s_{\text{cross}}=\mathrm{Acc}(B\!\to\!A)$ evaluates the paired model $B$ on images generated by $A$. We report $s_{\Delta}=s_{\text{cross}}-s_{\text{umm}}$.}
\label{tab:swap_test}
\centering
\small
\setlength{\tabcolsep}{4.5pt}
\renewcommand{\arraystretch}{1.08}
\begin{tabular}{llrrrr}
\toprule
\textbf{Pair } & \textbf{Target} 
& $\bm{s_{\Delta}}$ 
& $\bm{s_{\text{base}}}$ 
& $\bm{s_{\text{umm}}}$ 
& $\bm{s_{\text{cross}}}$ \\
\midrule
Ovis $\leftrightarrow$ UniWorld 
& Ovis     & -0.82 & 86.36 & 82.32 & 81.50 \\
& UniWorld & +0.34 & 84.85 & 75.76 & 76.10 \\
\midrule
Show-o2 $\leftrightarrow$ BAGEL 
& Show-o2 & -2.77 & 84.34 & 76.77 & 74.00 \\
& BAGEL   & +3.04 & 83.33 & 76.76 & 79.80 \\
\midrule
OmniGen2 $\leftrightarrow$ Janus 
& OmniGen2 & -3.06 & 81.82 & 73.74 & 70.68 \\
& Janus    & +5.83 & 78.79 & 67.17 & 73.00 \\
\bottomrule
\end{tabular}
\end{table}

\noindent\textbf{Q4. \textit{Is SGU endogenous, and can models exploit shortcut signals?}}

\noindent\textbf{A4.}
SGU is model-internal: all stages are performed by the tested UMM itself under a fixed protocol, without external judge models or human evaluators. To prevent hidden state leakage, each stage is executed in an isolated stateless session, and only the explicit intermediate artifact is passed forward to support the subsequent stage.

To probe whether loop-generated visual outputs embed model-specific shortcut cues that disproportionately benefit self-evaluation, we perform an image-swap test on the previously constructed high-scoring MMBench subset. We first pair UMMs with similar direct VQA accuracy so their base competence is comparable. For each target model, we compare two evaluations under the same questions. In \textit{Self}, the target answers on images it generates itself, whereas in \textit{Cross}, its paired model answers on the target's generated images. As summarized in Table~\ref{tab:swap_test}, \textit{Cross} remains broadly comparable to \textit{Self} across pairs, with only minor and non-systematic differences, suggesting no consistent shortcut signals that materially influence SGU performance. 

\section{Conclusion, Limitations, and Future Work}

This paper introduces Self-Generative-Understanding (SGU), an endogenous closed-loop evaluation protocol designed for Unified Multimodal Models. By reusing standard VQA benchmarks, SGU evaluates whether a model can operate through its own understanding-and-generation process and complete tasks grounded in the original images, providing a complementary system-level view rather than replacing component-wise evaluations. Experiments across four benchmarks and six representative UMMs show that SGU captures system-level behavior that is not fully reflected by isolated evaluations, while additional analyses further explain how the integrated process affects the final score and reveals model-specific bottlenecks.

\noindent\textbf{Limitations and Future Work.}
SGU is a system-level evaluation signal, not a standalone theoretical criterion for UMM alignment. Its score naturally aggregates effects from multiple stages, so fine-grained failure attribution should be performed with stage-wise analyses and component-level metrics. The current implementation also uses text as an intermediate representation, which may introduce information bottlenecks for detail-intensive tasks and fine-grained visual evidence. Future work will explore richer intermediate representations, broader task instantiations, and adapting SGU into a training objective using closed-loop feedback.

\bibliographystyle{plainnat}
\bibliography{references}

%%%%%%%%%%%%%%%%%%%%%%%%%%%%%%%%%%%%%%%%%%%%%%%%%%%%%%%%%%%%

\appendix

\section{Datasets and Evaluation Benchmarks}
\label{app:datasets}

We evaluate SGU on four representative multimodal benchmarks covering general perception, visual reasoning, mathematical reasoning, and text-centric visual understanding. These datasets provide diverse VQA-style tasks for examining the system-level behavior of UMMs under the SGU protocol.

\subsection{Dataset Descriptions}
\label{app:dataset_descriptions}

\textbf{MMStar}~\cite{MMStar}.
MMStar is a high-quality vision-language benchmark curated to emphasize \emph{vision-dependent} questions. It is designed to reduce the influence of language priors by filtering samples that can be answered reliably without the image, and covers diverse perception-centric skills such as fine-grained recognition, spatial relations, and multi-object reasoning.

\textbf{MMBench}~\cite{MMBench}.
MMBench is a broad multimodal evaluation suite targeting general visual perception and reasoning. Its questions span recognition and attributes, spatial understanding, commonsense and social reasoning, and multi-step visual reasoning, providing a comprehensive testbed for overall VQA competence under a unified format.

\textbf{MathVista}~\cite{MathVista}.
MathVista focuses on visually grounded mathematical reasoning, where solving requires jointly interpreting visual evidence, such as geometry diagrams, plots, tables, and charts, and performing mathematical inference. This benchmark is sensitive to errors in visual reconstruction, since small distortions in the reconstructed image may affect the evidence needed for correct reasoning.

\textbf{OCR-VQA}~\cite{OCR-VQA}.
OCR-VQA evaluates reading and reasoning over text embedded in images, including scene text and document-style content. Questions often require precise recognition of characters or words and their context, making it useful for analyzing failures involving text understanding and visual text reconstruction.

\subsection{Summary of Data Usage}
\label{app:data_usage}

Table~\ref{tab:dataset_summary} summarizes the splits and primary evaluation focus for each dataset used in SGU.

\begin{table}[t]
\caption{Dataset splits and primary evaluation focus used in this work.}
\label{tab:dataset_summary}
\centering
\small
\setlength{\tabcolsep}{4.2pt}
\renewcommand{\arraystretch}{1.06}
\begin{tabular}{lcll}
\toprule
\textbf{Dataset} & \textbf{Split} & \textbf{Type} & \textbf{Primary focus} \\
\midrule
\textbf{MMStar}    & Val       & General        & Vision-dependent reasoning; perception \\
\textbf{MMBench}   & Dev       & General        & Visual perception; general reasoning \\
\textbf{MathVista} & Test-mini & Math \& charts & Geometric structure; mathematical reasoning \\
\textbf{OCR-VQA}   & Val       & Text-in-image  & OCR; visual text understanding \\
\bottomrule
\end{tabular}
\end{table}

\section{Implementation Details}
\label{app:implementation}

This section provides the inference configurations and prompts used to execute the SGU protocol. All experiments are inference-only evaluations without model training or fine-tuning. We run the SGU pipeline on NVIDIA A800 GPUs with 80GB memory; the exact runtime varies across models and datasets because each sample requires captioning, image generation, and final VQA inference. More expensive ablations are conducted on fixed-size subsets to reduce compute while keeping controlled comparisons.

\subsection{Model Inference Configurations}
\label{app:inference_config}

To ensure fair comparison across UMMs, we use a consistent inference protocol whenever possible, while following each model's native generation interface.

\begin{itemize}
    \item \textbf{Understanding phase ($I \rightarrow T$).}
    For the image-to-text stage, we use deterministic decoding by setting \texttt{do\_sample} to \textbf{False}. The output length is capped with \texttt{max\_new\_tokens} set to \textbf{256}, which provides a unified resource constraint and standardizes the textual input to the subsequent generation stage.

    \item \textbf{Generation phase ($T \rightarrow I$).}
    Generation configurations differ across model architectures. For reproducibility, we fix the global random seed to \textbf{42}. When applicable, we use a Classifier-Free Guidance (CFG) scale of \textbf{5.0}. Specific settings follow each model's native architecture:
    \begin{itemize}
        \item \textit{Autoregressive architecture (Janus-Pro-7B):}
        We generate images at a resolution of $\mathbf{512 \times 512}$. The generation process uses \textbf{576 image tokens} with multinomial sampling and a temperature of 1.0.

        \item \textit{Diffusion and flow-based architectures (BAGEL, OmniGen2, Ovis, UniWorld, Show-o2):}
        We use \textbf{30 inference steps} and generate images at $\mathbf{512 \times 512}$ resolution. When supported by the model, the CFG scale is set to \textbf{5.0}.
    \end{itemize}

    \item \textbf{VQA phase.}
    For the final question-answering stage, we use deterministic decoding by setting \texttt{do\_sample} to \textbf{False}. The output length is restricted with \texttt{max\_new\_tokens} set to \textbf{64}, since VQA answers are typically concise.
\end{itemize}

\subsection{Prompt Engineering}
\label{app:prompts}

The prompts used in each phase of the SGU cycle are listed below. The placeholders \texttt{\{description\}}, \texttt{\{Question\}}, and \texttt{\{Options\}} denote the generated caption, the input question, and the candidate answer options, respectively.

\paragraph{Standard protocol.}
For the main experiments, we use the following default prompts:

\begin{itemize}
    \item \textbf{Phase 1: Understanding ($I \rightarrow T$)}
    \begin{quote}
        \texttt{<image> Describe this image in detail.}
    \end{quote}

    \item \textbf{Phase 2: Generation ($T \rightarrow I$)}
    \begin{quote}
        \texttt{Generate an image based on the following description: \{description\}.}
    \end{quote}

    \item \textbf{Phase 3: Evaluation (VQA)}
    \begin{itemize}
        \item \textit{Multiple-choice questions:}
        \begin{quote}
            \texttt{\{Question\} \{Options\} Answer with the option letter directly.}
        \end{quote}

        \item \textit{Open-form questions:}
        \begin{quote}
            \texttt{\{Question\} Answer the question directly.}
        \end{quote}
    \end{itemize}
\end{itemize}

\paragraph{Prompt variants for sensitivity analysis.}
For the prompt-sensitivity experiment, we vary one stage at a time while keeping the rest of the SGU pipeline unchanged. The prompt variants are:

\begin{itemize}
    \item \textbf{Understanding variants ($I \rightarrow T$):}
    \begin{itemize}
        \item \textit{Var-1 (Und.):}
        \begin{quote}
             \texttt{Describe the visual content of this image in detail.}
        \end{quote}

        \item \textit{Var-2 (Und.):}
        \begin{quote}
             \texttt{Analyze the image and provide a thorough description capturing all key elements.}
        \end{quote}
    \end{itemize}

    \item \textbf{Generation variants ($T \rightarrow I$):}
    \begin{itemize}
        \item \textit{Var-1 (Gen.):}
        \begin{quote}
             \texttt{\{description\}}
        \end{quote}

        \item \textit{Var-2 (Gen.):}
        \begin{quote}
             \texttt{Generate a high-quality, detailed image of \{description\}.}
        \end{quote}
    \end{itemize}
\end{itemize}

\subsection{Stateless Pipeline Execution}
\label{app:stateless}

SGU is executed in a \emph{stateless} manner across stages. The captioning, generation, and final VQA steps are run as independent inference calls. Only the explicit intermediate artifact, i.e., the generated caption or reconstructed image, is passed to the next stage; no hidden states, KV-cache, or conversation history are carried over.

\paragraph{Stage isolation.}
For each stage, we create a fresh model invocation with a clean input payload. We do not reuse the same chat session or inference context across stages, and we avoid accumulating messages in a multi-turn dialog. This ensures that the only information available at each step is what is explicitly provided in that step's input.

\paragraph{Inputs to each stage.}
(i) \textbf{Captioning} takes the original image $v$ and the captioning prompt, and outputs a description $t_g$.
(ii) \textbf{Generation} takes only the text prompt formed from $t_g$ and produces a reconstructed image $\hat{v}$.
(iii) \textbf{VQA scoring} takes only $\hat{v}$ and the question $q$, and returns the answer used to compute $s_{\text{umm}}$; the original image $v$ and the intermediate caption $t_g$ are not provided to the final VQA stage.

\section{Additional Metrics and Theoretical Details}
\label{app:metrics_theory}

\subsection{Auxiliary Metrics for Isolated Assessment}
\label{app:aux_metrics}

In addition to the end-to-end SGU score, we report several commonly used isolated metrics to characterize individual components. These metrics are not used as the SGU objective, but serve as auxiliary references for analysis.

\paragraph{FID (Fr\'echet Inception Distance)}~\cite{heusel2017ttur}.
FID measures the distributional distance between two image sets by comparing their feature statistics in a pretrained Inception network. Given feature activations with empirical means and covariances $(\mu_r,\Sigma_r)$ for real images and $(\mu_g,\Sigma_g)$ for generated images, FID is defined as
\begin{equation}
\mathrm{FID}
=
\|\mu_r-\mu_g\|_2^2
+
\mathrm{Tr}\!\left(
\Sigma_r+\Sigma_g-2(\Sigma_r\Sigma_g)^{1/2}
\right),
\end{equation}
where $\mathrm{Tr}(\cdot)$ denotes the matrix trace.

\paragraph{CLIPScore}~\cite{hessel2022clipscorereferencefreeevaluationmetric}.
We use CLIP embeddings to quantify (i) visual similarity between the original and reconstructed images, and (ii) text--image similarity between the model-generated caption and the original image. Let $f_I(\cdot)$ and $f_T(\cdot)$ denote the CLIP image and text encoders, and let $\langle \cdot,\cdot\rangle$ denote cosine similarity. We compute
\begin{align}
\mathrm{CLIP\text{-}I}
&=
\left\langle
\frac{f_I(I)}{\|f_I(I)\|},
\frac{f_I(I')}{\|f_I(I')\|}
\right\rangle, \\
\mathrm{CLIP\text{-}T}
&=
\left\langle
\frac{f_I(I)}{\|f_I(I)\|},
\frac{f_T(C)}{\|f_T(C)\|}
\right\rangle,
\end{align}
where $I$ is the original image, $I'$ is the reconstructed image, and $C$ is the intermediate caption produced at the understanding stage.

\subsection{Direct-VQA Reference for SGU}
\label{app:upper_bound}

We use direct VQA accuracy on the original image as a model-specific upper-bound reference for interpreting the SGU score:
\begin{equation}
s_{\text{base}}
=
\mathbb{E}_{(v,q,a)\sim\mathcal{D}}
\left[
\mathbb{I}\left(
\operatorname{Match}\big(\mathcal{M}_U(v,q),a\big)
\right)
\right].
\end{equation}
Compared with direct VQA, SGU introduces an additional understanding--generation loop before the final answer is produced. The model must first convert the original image into an intermediate textual representation, reconstruct a visual context from it, and then answer the original question using the reconstructed image. Therefore, under the intended semantic-preservation setting, $s_{\text{base}}$ serves as a natural reference for the best performance the model can achieve without the additional loop.

\paragraph{Proposition (upper-bound reference under semantic preservation).}
Assume that the SGU loop does not introduce new task-solving semantic evidence beyond what is extracted from the original image by the model's understanding pathway. Under this assumption, direct VQA accuracy on the original image serves as an upper-bound reference for the SGU score:
\begin{equation}
    s_{\text{umm}} \leq s_{\text{base}}.
\end{equation}

\paragraph{Proof.}
We provide an abstraction-level argument based on semantic information preservation.

\noindent\textbf{Semantic abstraction.}
Let $m_U(v)$ denote the semantic information extracted from image $v$ by the visual understanding pathway of the unified model $\mathcal{M}$, viewed as a mapping from the visual modality to a semantic space. Similarly, let $m_G(t)$ denote the semantic information realized in the generated image from text $t$ by the generation pathway of the model.

For a data sample $d=(v,q,a)\in\mathcal{D}$, let $\mathcal{H}_v$ denote the complete semantic space contained in image $v$. We define the minimal semantic set required to correctly answer the query as
\begin{equation}
    \mathcal{B}_d
    \coloneqq
    \Big\{
        s \subseteq \mathcal{H}_v
        \;\big|\;
        \operatorname{Match}\big(\mathcal{M}_U(s,q),a\big)=1
    \Big\}.
\end{equation}
By construction, $\mathcal{B}_d \subseteq \mathcal{H}_v$.

\noindent\textbf{Correctness sets.}
We define the sets of correctly answered samples under direct VQA and SGU as
\begin{align}
    \mathcal{G}_{\text{base}}
    &\coloneqq
    \Big\{
        (v,q,a)\in\mathcal{D}
        \;\big|\;
        \operatorname{Match}\big(\mathcal{M}_U(v,q),a\big)=1
    \Big\}, \\
    \mathcal{G}_{\text{SGU}}
    &\coloneqq
    \Big\{
        (v,q,a)\in\mathcal{D}
        \;\big|\;
        \operatorname{Match}\big(
        \mathcal{M}_U(\mathcal{M}_G(\mathcal{M}_U(v)),q),a
        \big)=1
    \Big\}.
\end{align}
To show $s_{\text{umm}}\leq s_{\text{base}}$, it suffices to show
\begin{equation}
    \mathcal{G}_{\text{SGU}} \subseteq \mathcal{G}_{\text{base}}.
\end{equation}

Consider any sample $d^*=(v^*,q^*,a^*)\in\mathcal{G}_{\text{SGU}}$. By definition, the semantic information preserved after the understanding--generation composition,
\begin{equation}
    m_G \circ m_U(v^*),
\end{equation}
is sufficient to contain the semantic evidence $\mathcal{B}_{d^*}$ required to answer $q^*$ correctly.

Under the semantic-preservation assumption, the generation process cannot introduce new task-solving semantic evidence beyond what is encoded in $m_U(v^*)$. Therefore,
\begin{equation}
    \big( m_G \circ m_U(v^*) \big)
    \cap \mathcal{B}_{d^*}
    \subseteq
    m_U(v^*) \cap \mathcal{B}_{d^*}.
\end{equation}
Since $d^*\in\mathcal{G}_{\text{SGU}}$, the SGU loop preserves sufficient information for the correct answer, which implies
\begin{equation}
    \mathcal{B}_{d^*}
    =
    \big( m_G \circ m_U(v^*) \big) \cap \mathcal{B}_{d^*}.
\end{equation}
Combining the two relations gives
\begin{equation}
    \mathcal{B}_{d^*}
    \subseteq
    m_U(v^*) \cap \mathcal{B}_{d^*}
    \subseteq
    \mathcal{B}_{d^*}.
\end{equation}
Thus,
\begin{equation}
    \mathcal{B}_{d^*}
    =
    m_U(v^*) \cap \mathcal{B}_{d^*},
\end{equation}
which implies
\begin{equation}
    \mathcal{B}_{d^*} \subseteq m_U(v^*).
\end{equation}
Therefore, the original visual understanding representation $m_U(v^*)$ already contains sufficient semantic information to answer $q^*$ correctly, implying $d^*\in\mathcal{G}_{\text{base}}$.

Since $d^*$ is arbitrary, we conclude
\begin{equation}
    \mathcal{G}_{\text{SGU}} \subseteq \mathcal{G}_{\text{base}},
\end{equation}
and therefore
\begin{equation}
    s_{\text{umm}} \leq s_{\text{base}}.
\end{equation}
\hfill$\square$

The above argument provides a semantic-space interpretation of why $s_{\text{base}}$ is a useful upper-bound reference for SGU. In practice, rare cases may occur where the reconstructed image makes a previously incorrect direct-VQA sample easier to answer. Therefore, we use $s_{\text{base}}$ as a model-specific reference rather than as an unconditional guarantee for every individual sample.

\begin{figure}[t]
    \centering
    \includegraphics[width=1\linewidth]{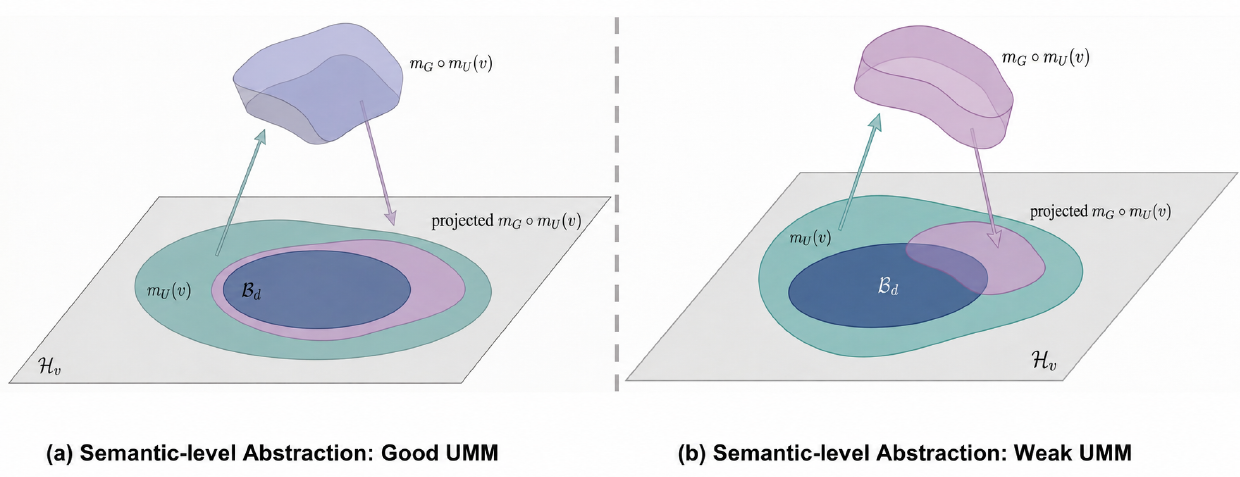}
    \caption{Abstraction-level illustration of SGU success and failure.
    (a) The understanding--generation composition retains the information needed for the target question.
    (b) Part of the information is lost during the loop, causing the reconstructed image to no longer support the correct answer.}
    \label{fig:abstraction}
\end{figure}

Figure~\ref{fig:abstraction} visualizes this abstraction. In case (a), the understanding--generation composition retains the information needed to answer the question, so the model remains correct after the SGU loop. In case (b), part of the relevant visual information is lost during the loop, leading to an incorrect final answer. This illustrates how SGU reflects the integrated behavior of understanding, intermediate representation, and generation.

\section{More SGU Case Examples}
\label{app:cases}

\begin{figure}[t]
    \centering
    \includegraphics[width=\linewidth]{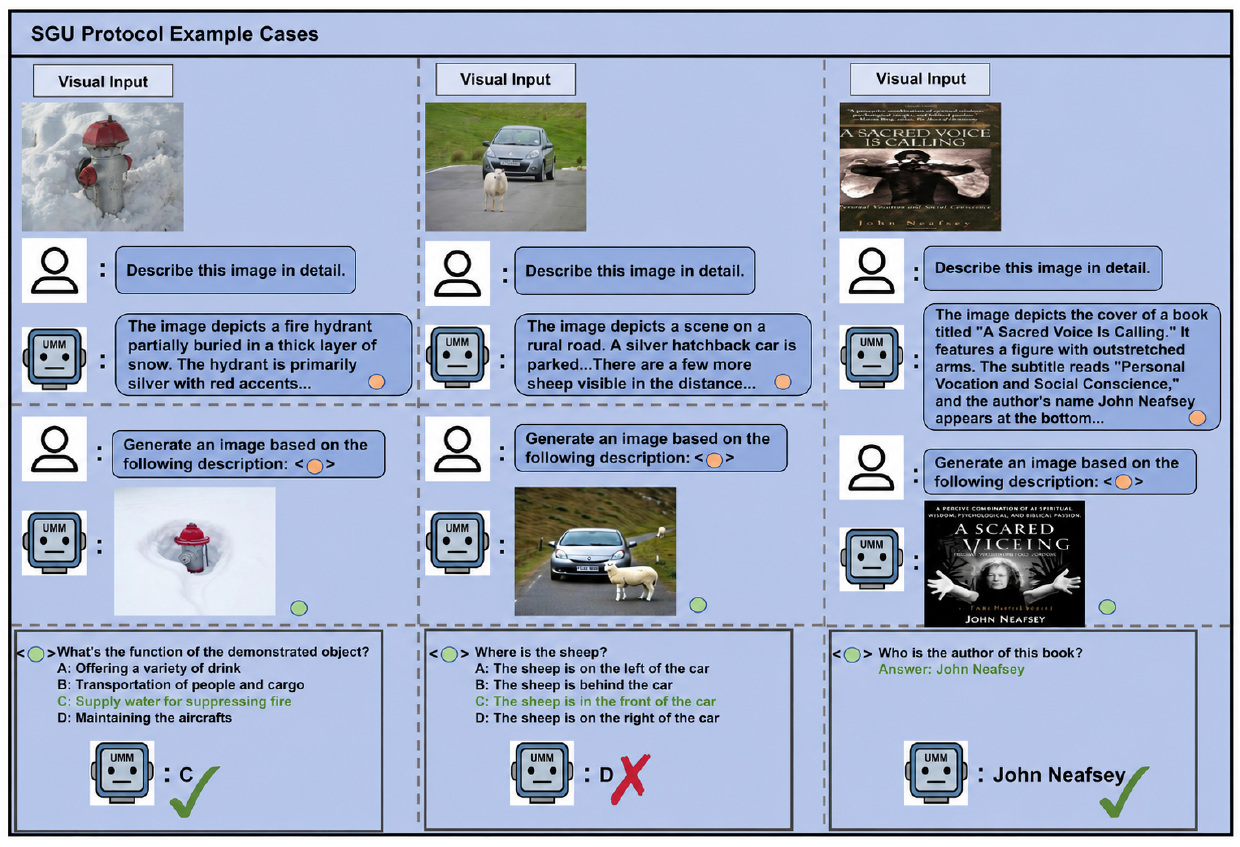}
    \caption{Additional SGU case examples showing the complete loop from the input VQA sample to the intermediate caption, reconstructed image, and final answer.}
    \label{fig:supp_captioner}
\end{figure}

Figure~\ref{fig:supp_captioner} provides additional qualitative examples of the full SGU loop. These cases illustrate that the final grounded task can be answered correctly only when the model performs well across the integrated process, including image understanding, intermediate description, visual generation, and final reasoning. Weaknesses in any stage may affect the final answer and are therefore reflected in the SGU evaluation result.

\section{Subset Construction for Ablations}
\label{app:subset}

Our ablation studies require repeatedly running the full SGU loop, including captioning, image generation, and self-VQA, under controlled modifications. This is substantially more expensive than isolated evaluation. We therefore construct fixed-size subsets for ablations. These studies do not aim to estimate absolute benchmark performance; instead, they analyze \emph{relative} changes induced by interventions, such as stage-wise replacement or prompt variation. To ensure comparability, all ablation settings are evaluated on the same fixed subset split, and each subset is first evaluated with the standard SGU protocol as the baseline before applying modifications.

\paragraph{Stratified sampling.}
We build each subset via stratified random sampling with a fixed seed (42), preserving the type composition of the corresponding full split when such labels are available. For MMStar and MMBench, we stratify by the provided category labels. For MathVista, which does not provide a unified fine-grained category taxonomy in our setup, we stratify by question format, i.e., multiple-choice versus non-multiple-choice. For OCR-VQA, where questions are open-ended and no category labels are available, we uniformly sample from the split.

\paragraph{Subset specifications.}
Table~\ref{tab:subset_specs} summarizes the subset size and stratification granularity used for each dataset. For MMStar and MMBench, we use $198$ instances and preserve category distribution; for MMStar this yields a balanced subset across its six categories (33 each). For MathVista, we sample $200$ instances while matching the MCP/non-MCP ratio (54\% vs.\ 46\%). For OCR-VQA, we uniformly sample $200$ instances using the same seed. These constructed subsets are used only for controlled ablations, including stage-wise replacement, prompt sensitivity, and shortcut-cue checks. All main results are reported on the full evaluation splits, while the subsets provide an efficient and consistent testbed for analyzing how specific factors affect SGU within the complete loop.

\paragraph{Representativeness check.}
To verify that these fixed-size subsets are suitable for ablations, we also run the full SGU protocol on each constructed subset and report the results in Table~\ref{tab:subset_results}. We observe that the subset-based results broadly follow the same trends as the full-benchmark results in Table~\ref{tab:main_results}, including the overall model ordering and the score distribution across datasets. This indicates that the subsets preserve the key evaluation behaviors of the original benchmarks and provide a consistent basis for controlled ablation analyses.

\begin{table}[t]
\caption{Ablation subsets constructed with a fixed random seed (42). For category-labeled datasets, we stratify by dataset categories; for MathVista, we stratify by question format; for OCR-VQA, we uniformly sample due to missing type labels.}
\label{tab:subset_specs}
\centering
\small
\setlength{\tabcolsep}{4.2pt}
\renewcommand{\arraystretch}{1.08}
\begin{tabular}{lccc}
\toprule
\textbf{Dataset} & \textbf{Subset size $N$} & \textbf{Stratification signal} & \textbf{Preserved composition} \\
\midrule
MMStar    & 198 & Category label  & 6 categories, 33 each \\
MMBench   & 198 & Category label  & 20 categories, proportional \\
MathVista & 200 & Question format & MCP 54\% / non-MCP 46\% \\
OCR-VQA   & 200 & N/A             & Uniform sample \\
\bottomrule
\end{tabular}
\end{table}

\begin{table*}[t]
\caption{SGU results on the constructed fixed-size subsets. For each dataset, we report direct VQA accuracy on the original image ($s_{\text{base}}$) and SGU score ($s_{\text{umm}}$).}
\label{tab:subset_results}
\centering
\small
\setlength{\tabcolsep}{4.2pt}
\renewcommand{\arraystretch}{1.10}
\begin{tabular}{l *{10}{c}}
\toprule
\multirow{3}{*}{\textbf{Model}}
& \multicolumn{2}{c}{\textbf{MMStar}}
& \multicolumn{2}{c}{\textbf{MMBench}}
& \multicolumn{2}{c}{\textbf{MathVista}}
& \multicolumn{2}{c}{\textbf{OCR-VQA}}
& \multicolumn{2}{c}{\textbf{Avg}} \\
\cmidrule(lr){2-3}
\cmidrule(lr){4-5}
\cmidrule(lr){6-7}
\cmidrule(lr){8-9}
\cmidrule(lr){10-11}
& Original & SGU
& Original & SGU
& Original & SGU
& Original & SGU
& Original & SGU \\
& $s_{\text{base}}$ & $s_{\text{umm}}$
& $s_{\text{base}}$ & $s_{\text{umm}}$
& $s_{\text{base}}$ & $s_{\text{umm}}$
& $s_{\text{base}}$ & $s_{\text{umm}}$
& $s_{\text{base}}$ & $s_{\text{umm}}$ \\
\midrule
\textsc{Upper-bound Ref.}
& \multicolumn{2}{c}{64.65}
& \multicolumn{2}{c}{86.36}
& \multicolumn{2}{c}{69.19}
& \multicolumn{2}{c}{82.32}
& \multicolumn{2}{c}{75.63} \\
\midrule

Janus-Pro-7B
& 41.92 & 33.84
& 78.79 & 67.17
& 41.92 & 33.84
& 68.69 & 36.87
& 57.83 & 42.93 \\

UniWorld-V1
& 56.06 & 37.37
& 84.85 & 75.76
& 63.64 & 38.38
& 79.29 & 30.30
& 70.96 & 45.45 \\

Show-o2-7B
& 53.03 & \underline{42.93}
& 84.34 & \underline{76.77}
& 47.98 & \underline{41.41}
& 63.13 & 32.32
& 62.12 & 48.36 \\

Ovis-U1-3B
& 60.61 & \textbf{44.95}
& 86.36 & \textbf{82.32}
& 68.69 & \textbf{45.45}
& 78.28 & 33.84
& 73.48 & 51.64 \\

BAGEL-7B
& 64.65 & 40.40
& 83.33 & \underline{76.77}
& 69.19 & \underline{41.41}
& 72.22 & \underline{49.49}
& 72.35 & \underline{52.02} \\

OmniGen2
& 50.51 & 38.89
& 81.82 & 73.74
& 59.60 & 38.89
& 82.32 & \textbf{57.07}
& 68.56 & \textbf{52.15} \\

\bottomrule
\end{tabular}
\end{table*}

\section{Additional Experimental Results}
\label{app:additional_results}

This section reports supplementary quantitative results omitted from the main paper for brevity. We include full results of the stage-wise replacement study on the MathVista and OCR-VQA subsets, as well as intermediate CLIP-based signals collected on the full benchmark evaluation.

\subsection{Full Results for Stage-wise Replacement Study}
\label{app:replacement_full}

Table~\ref{tab:stage_replace_mathvista_full} and Table~\ref{tab:stage_replace_ocrvqa_full} provide the complete stage-wise replacement results on the constructed MathVista and OCR-VQA subsets. For each model, we report the baseline SGU score and the SGU score after replacing either the captioning, i.e., image-to-text understanding, stage or the image-generation stage, while keeping all other stages unchanged.

\begin{table}[t]
\centering
\begin{minipage}[t]{0.48\linewidth}
\centering
\captionof{table}{Stage-wise replacement on the constructed \textbf{MathVista} subset. We report the baseline SGU score and the score after replacing either the captioning or generation stage.}
\label{tab:stage_replace_mathvista_full}
\small
\setlength{\tabcolsep}{4.2pt}
\renewcommand{\arraystretch}{1.08}
\begin{tabular}{lccc}
\toprule
\textbf{Model} & \textbf{Base} & \textbf{Cap.} & \textbf{Gen.} \\
\midrule
UniWorld-V1  & 38.38 & 34.34 & 52.02 \\
Janus-Pro-7B & 33.84 & 36.87 & 37.88 \\
Show-o2-7B   & 41.41 & 39.39 & 41.41 \\
Ovis-U1-3B   & 45.45 & 46.46 & 52.02 \\
BAGEL-7B     & 41.41 & 39.90 & 46.46 \\
OmniGen2     & 38.89 & 42.93 & 50.51 \\
\bottomrule
\end{tabular}
\end{minipage}
\hfill
\begin{minipage}[t]{0.48\linewidth}
\centering
\captionof{table}{Stage-wise replacement on the constructed \textbf{OCR-VQA} subset. We report the baseline SGU score and the score after replacing either the captioning or generation stage.}
\label{tab:stage_replace_ocrvqa_full}
\small
\setlength{\tabcolsep}{4.2pt}
\renewcommand{\arraystretch}{1.08}
\begin{tabular}{lccc}
\toprule
\textbf{Model} & \textbf{Base} & \textbf{Cap.} & \textbf{Gen.} \\
\midrule
UniWorld-V1  & 30.30 & 27.78 & 69.19 \\
Janus-Pro-7B & 36.87 & 33.84 & 58.08 \\
Show-o2-7B   & 32.32 & 33.33 & 61.62 \\
Ovis-U1-3B   & 33.84 & 35.86 & 68.69 \\
BAGEL-7B     & 49.49 & 48.99 & 62.63 \\
OmniGen2     & 57.07 & 55.56 & 69.19 \\
\bottomrule
\end{tabular}
\end{minipage}
\end{table}

\subsection{Intermediate CLIP-T and CLIP-I Signals on Full Benchmarks}
\label{app:clip_signals}

To connect SGU with conventional component-level evaluation signals, we additionally report intermediate similarity metrics. Specifically, we compute \textbf{CLIP-T} between the model-generated caption and the original image, and \textbf{CLIP-I} between the reconstructed image and the original image, both on the full evaluation sets. These metrics are not used to compute SGU, but serve as auxiliary references for interpreting the captioning and reconstruction stages.

\begin{table}[t]
\caption{CLIP-T similarity between each model-generated caption and the original image across the full evaluation datasets. Higher is better.}
\label{tab:clip_t_full}
\centering
\small
\setlength{\tabcolsep}{8pt}
\renewcommand{\arraystretch}{1.08}
\begin{tabular}{lcccc}
\toprule
\textbf{Model} & \textbf{MMStar} & \textbf{MMBench} & \textbf{MathVista} & \textbf{OCR-VQA} \\
\midrule
Janus-Pro-7B & 0.2841 & 0.2765 & 0.3102 & 0.3499 \\
BAGEL-7B     & 0.2817 & 0.2703 & 0.3081 & 0.3158 \\
UniWorld-V1  & 0.2820 & 0.2749 & 0.3026 & 0.3555 \\
Show-o2-7B   & 0.2790 & 0.2704 & 0.3050 & 0.3189 \\
OmniGen2     & 0.2794 & 0.2710 & 0.3038 & 0.3515 \\
Ovis-U1-3B   & 0.2822 & 0.2778 & 0.3040 & 0.3497 \\
\bottomrule
\end{tabular}
\end{table}

\begin{table}[!t]
\caption{CLIP-I similarity between the reconstructed image and the original image across the full evaluation datasets. Higher is better.}
\label{tab:clip_i_full}
\centering
\small
\setlength{\tabcolsep}{8pt}
\renewcommand{\arraystretch}{1.08}
\begin{tabular}{lcccc}
\toprule
\textbf{Model} & \textbf{MMStar} & \textbf{MMBench} & \textbf{MathVista} & \textbf{OCR-VQA} \\
\midrule
Janus-Pro-7B & 0.7126 & 0.7736 & 0.6494 & 0.6373 \\
BAGEL-7B     & 0.7213 & 0.7577 & 0.6873 & 0.6881 \\
UniWorld-V1  & 0.5790 & 0.6849 & 0.5058 & 0.3454 \\
Show-o2-7B   & 0.6712 & 0.7316 & 0.6340 & 0.4752 \\
OmniGen2     & 0.6946 & 0.7430 & 0.6860 & 0.6894 \\
Ovis-U1-3B   & 0.7042 & 0.7515 & 0.6831 & 0.4909 \\
\bottomrule
\end{tabular}
\end{table}

\section{Additional Discussion on Robustness, Information Bottlenecks, and Score Interpretation}
\label{app:discussion}

This section provides additional discussion on robustness, information bottlenecks, and score interpretation in the SGU protocol. We focus on three issues: randomness in the evaluation pipeline, the role of the intermediate textual representation, and whether SGU is dominated by a single stage such as captioning or generation.

\paragraph{Random seeds and evaluation robustness.}
In our experiments, we use consistent random seeds across all models to ensure fair comparison under the same evaluation protocol. We do not explicitly report a full random-seed sweep, since running the complete SGU loop is computationally expensive. Instead, our prompt-sensitivity analysis introduces structured perturbations to the framework by varying the understanding and generation prompts while keeping the rest of the pipeline unchanged. As shown in Table~\ref{tab:prompt_sens}, the results remain close to the default setting under these perturbations, suggesting that SGU provides a stable signal under moderate implementation changes, including stochastic factors controlled by the fixed inference setup.

\begin{table}[!htbp]
\caption{Text-bottleneck analyses on OCR-VQA. Left: caption-length sensitivity under caption-only QA with Janus-Pro-7B. Right: caption-only QA comparison with OmniGen2.}
\label{tab:text_bottleneck_analysis}
\centering
\footnotesize
\setlength{\tabcolsep}{6pt}
\renewcommand{\arraystretch}{0.95}
\begin{tabular}{lc|lc}
\toprule
\multicolumn{2}{c|}{\textbf{Caption Length}} 
& \multicolumn{2}{c}{\textbf{Caption-only QA}} \\
\cmidrule(lr){1-2}\cmidrule(lr){3-4}
\textbf{Setting} & \textbf{Acc.} 
& \textbf{Setting} & \textbf{Acc.} \\
\midrule
Janus-Pro-7B, 256 tokens & 49.50 
& Original image QA ($s_{\text{base}}$) & 79.39 \\
Janus-Pro-7B, 512 tokens & 49.53 
& Caption-only QA & 60.05 \\
-- & -- 
& Full SGU ($s_{\text{umm}}$) & 56.59 \\
\bottomrule
\end{tabular}
\end{table}

\paragraph{Intermediate text bottleneck.}
The intermediate caption in SGU is a compact textual representation of the model's visual understanding, rather than a lossless encoding of the image. This bottleneck should be interpreted as part of the evaluated process. If the model fails to extract or express important visual information in the captioning stage, this reflects limitations in its understanding and representation ability under the SGU protocol. In our default setting, the model is explicitly instructed to describe the image in detail. Therefore, missing information in the generated caption is not only an artifact of the framework, but also an informative signal of how well the model can form a general visual representation.

One possible alternative is to condition the captioning stage on the downstream question. While this may reduce information loss for a specific query, it changes the nature of the evaluation: the intermediate representation becomes question-specific rather than a general description of the image. SGU instead adopts a question-agnostic captioning stage to evaluate whether a UMM can first form a general visual description and then use its own generated content to support downstream reasoning. This design better matches our goal of evaluating the model as an integrated system, rather than optimizing the intermediate caption for a particular question.

\paragraph{Caption length.}
To examine whether the text bottleneck is mainly caused by the caption token limit, we conduct a caption-only QA analysis with different maximum caption lengths. As shown in Table~\ref{tab:text_bottleneck_analysis}, increasing the caption limit from 256 to 512 tokens on OCR-VQA with Janus-Pro-7B changes caption-only QA accuracy only marginally, from $49.50$ to $49.53$. This suggests that, under our current setting, the observed degradation is not simply determined by the 256-token limit. Instead, it also depends on how effectively the model understands, organizes, and expresses visual information in the intermediate representation.

\paragraph{Understanding and generation bottlenecks.}
SGU should not be interpreted as being determined by a single bottleneck. Stage-wise replacement results in Section~\ref{app:replacement_full} show that replacing the generation stage often leads to larger improvements, indicating that visual reconstruction is an important bottleneck for current UMMs. However, this does not mean that SGU only measures generation. If the captioning stage fails to extract sufficient visual information, a stronger generator cannot recover information that is absent from the intermediate representation. Conversely, even a good caption may still lead to performance degradation if the model fails to reconstruct a faithful visual context or reason correctly over the reconstructed image.

To further examine the role of the understanding-side intermediate representation, we perform caption-only QA, where the model answers the question directly from its generated caption without the image-generation stage. As shown in Table~\ref{tab:text_bottleneck_analysis}, on OCR-VQA with OmniGen2, direct VQA on the original image achieves $79.39$, caption-only QA obtains $60.05$, and the full SGU loop obtains $56.59$. The drop from original-image QA to caption-only QA shows that the captioning and reasoning path already introduces substantial degradation, while the further drop under full SGU reflects additional challenges from visual generation and reconstruction. These results support our view that SGU evaluates the combined effects of understanding, intermediate representation, generation, and final reasoning, rather than serving as a metric for any single component alone.

\paragraph{Interpretation of the SGU score.}
The SGU score is intended as an outcome-based system-level evaluation signal. It does not replace component-wise understanding or generation metrics, nor does it independently attribute errors to a specific stage. Instead, it evaluates whether the UMM can complete the full closed-loop process using its own outputs. Component-level metrics and stage-wise analyses remain useful for diagnosis, while SGU provides a complementary view of whether the integrated understanding-and-generation process succeeds on grounded downstream tasks.

%%%%%%%%%%%%%%%%%%%%%%%%%%%%%%%%%%%%%%%%%%%%%%%%%%%%%%%%%%%%

\end{document}